\PassOptionsToPackage{numbers,compress}{natbib}
\documentclass{article}
\usepackage[preprint]{neurips_2026}
\newif\ifextended\extendedfalse
\usepackage[utf8]{inputenc}
\usepackage[T1]{fontenc}
\usepackage{amsmath,amssymb,booktabs,graphicx,float,microtype,url}
\usepackage[hidelinks]{hyperref}
\title{Concept Direction Reliability Across Languages\\with Different Tokenizer Fertility}
\author{%
  Muhammad Abdullahi Said $^{\text{1,5}}$ \quad Abass Oguntade $^{\text{1}}$ \quad Elisha Komolafe $^{\text{1}}$ \\
  Babangida Sani $^{\text{2}}$ \quad Fatima Muhammad Adam $^{\text{3}}$ \quad Muhammad Sammani Sani $^{\text{4}}$ \\
  $^{1}$African Institute for Mathematical Sciences \quad $^{2}$Bayero University Kano \\
  $^{3}$Federal University Dutse \quad $^{4}$University of Vienna \quad $^{5}$University of Cape Town \\
  \texttt{mohdasaid@aims.ac.za}
}

\newcommand{\R}{\mathbb{R}}
\newcommand{\dir}{\widehat{d}}

\newcommand{\mainpage}{\clearpage}

\begin{document}
\maketitle
\begin{abstract}
Extracted sentiment directions can vary across samples even when downstream sentiment classification remains accurate. To evaluate direction reproducibility, we measure split-half agreement in English, Hausa, and Yoruba representations across four language models using both native and translated texts. We identify layers selected for agreement using ten topics and evaluate direction agreement across separate groups of fifteen topics. Using the final token, split-half agreement ranges from 0.737 to 0.870 for English, 0.589 to 0.762 for Hausa, and 0.101 to 0.399 for Yoruba, maintaining this language rank order across all 77 complete model comparisons. Classifiers trained on these same layers consistently predict sentiment above chance, demonstrating that predictive accuracy does not imply directional consistency. Furthermore, averaging token representations yields less consistent agreement, and high agreement can partially reflect sentence length. Ultimately, our findings highlight the need to measure vector direction reproducibility independently of classification performance, though they do not establish that tokenizer fertility which is the average number of tokens per whitespace separated word causes cross-lingual differences.
\end{abstract}

\section{Introduction}
Concept directions are widely used to identify and manipulate properties such as sentiment within language model representations \citep{turner2023actadd,zou2023repe}. A standard extraction approach computes the difference in means between representations of contrastive text pairs. However, before interpreting or steering with an extracted direction, it is essential to determine whether an independent sample would yield a similar vector.

We evaluate direction reproducibility for sentiment across English, Hausa, and Yoruba. Each contrast pair consists of a pleased message and an annoyed message regarding a shared topic. Our primary evaluation metric is split-half agreement: we repeatedly partition topics into two non-overlapping subsets, estimate directions from each, and measure their consistency across samples. Keeping each topic within one half prevents agreement from being inflated by sharing that topic across halves.

To contextualize our findings, we compare direction agreement against linear probe accuracy \citep{alain2016probes,belinkov2022probing}. While probe accuracy assesses whether sentiment information is linearly recoverable, direction agreement measures whether a specific extraction method reliably isolates the same vector. 

Finally, we analyze how direction agreement interacts with factors such as representation method, text source, sentence length, topic selection, and tokenizer fertility \citep{rust2021tokenizer,ahia2023cost}. We measure the association between fertility and agreement across languages without claiming a direct causal link.

\mainpage
\section{Study design}
\label{sec:method}
\paragraph{Data and models.}
Each language dataset comprises 100 sentiment contrast pairs across 25 topics (4 pairs per topic). While English contains source pairs only, Hausa and Yoruba are evaluated across four distinct text sources (arms): native texts written directly by local authors (Arm A), human translations of English pairs (Arm B), machine translations of English pairs (Arm C), and back-translated native texts (Arm D). Native writers were prompted in their own language without viewing the English sentences. We apply Unicode NFC normalization across all texts and evaluate each arm independently, as translation can alter the structural features used by a model \citep{artetxe2020artifacts}.

We evaluate representations across four open-weight models: Gemma 4 (E2B, E4B, and 12B variants) \citep{gemmateam2026} and AfroLlama V1 \citep{afrollama2024}, where E2B and E4B denote effective active parameter counts instead of total stored weights. At every layer, we extract representations using both the final token position and mean pooling across all non-padding tokens. All inference uses unquantized float16 precision with sequence lengths capped at 128 tokens; complete model checkpoints and extraction settings are provided in Appendix~\ref{app:protocol}.

\paragraph{Split-half agreement and layer selection.}
Let $h_i^+,h_i^-\in\R^p$ be the representations of the positive and negative messages in pair $i$, and let $\Delta_i=h_i^+-h_i^-$. For a set of pairs $S$, we normalise the average difference to obtain
\begin{equation}
\dir(S)=\frac{|S|^{-1}\sum_{i\in S}\Delta_i}
{\left\||S|^{-1}\sum_{i\in S}\Delta_i\right\|_2}.
\label{eq:direction}
\end{equation}
We measure agreement between two directions using cosine similarity, where $1$ indicates identical directions, $0$ indicates orthogonality, and negative values indicate opposing components. We use ten topics for layer selection and fifteen topics for evaluation. Within each topic set, we repeatedly partition the topics into two non-overlapping halves, estimate a direction from each half, and compute their cosine similarity. We average this similarity across 100 random topic partitions to determine the split-half agreement score, assessing whether independent samples recover consistent directions. Keeping all four contrast pairs of a given topic together within the same split prevents shared topics between the two halves. For evaluation, each split divides the fifteen topics into subsets of seven and eight topics (28 and 32 pairs, respectively). Given topic partitions $S_b$ and $\bar S_b$ for split $b$, the overall agreement $r$ is
\begin{equation}
r=\frac{1}{100}\sum_{b=1}^{100}\dir(S_b)^\top\dir(\bar S_b).
\label{eq:agreement}
\end{equation}
To check for sentence length effects, we average the two representations within each pair and construct a length direction by contrasting pairs above and below the median word count. We calculate length overlap as the absolute cosine similarity between this length direction and the sentiment direction. For a given model, we select the layer that maximizes selection agreement while maintaining a length overlap below $0.15$. If no layer satisfies this threshold, the selected layer is deemed ineligible and the result is reported as unavailable. This step addresses one possible influence of length, though it does not eliminate every source of bias.

\paragraph{Probes and uncertainty.}
We train logistic regression probes on the 40 selection pairs and test them on the 60 evaluation pairs. We report test accuracy at two layers: the layer selected for direction reliability, and a separate layer selected strictly via classification performance on selection data using 5-fold cross-validation that keeps each topic together. Feature vectors are centered and scaled to unit variance before training. We apply $L_2$ regularization with a fixed hyperparameter $C=1$ across all models. To estimate uncertainty while holding layer selection fixed, we construct confidence intervals across topics. Agreement intervals are computed using a leave-one-out jackknife procedure across evaluation topics. Accuracy intervals are constructed by bootstrapping the evaluation topics 1,000 times while keeping the fitted classifier fixed. These intervals do not adjust for multiple comparisons. Appendix~\ref{app:uncertainty} provides complete details for both estimation procedures.

\mainpage
\section{Sentiment prediction and direction agreement}
\label{sec:native}
\begin{figure}[H]
\centering
\includegraphics[width=\linewidth]{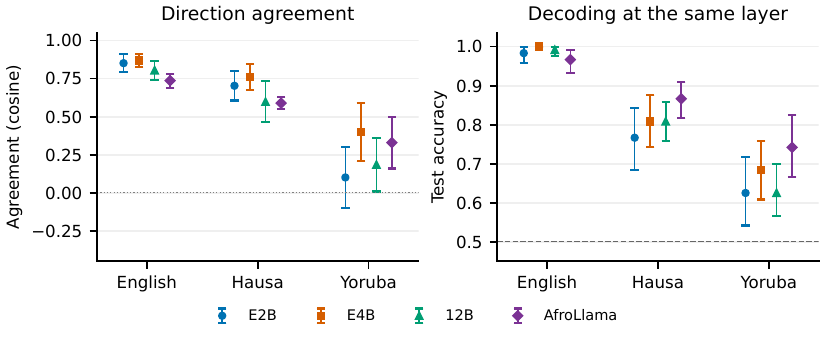}
\caption{Results for natively written texts in each language using final token representations. The panels show split-half direction agreement (left) and probe classification accuracy at the selected layer (right). Bars indicate approximate 95\% confidence intervals with layer selection held fixed. The dashed line marks chance accuracy.}
\label{fig:native}
\end{figure}

Using final token representations, split-half agreement is highest for English, followed by Hausa and then Yoruba across all four models (Figure~\ref{fig:native}), with values from 0.737 to 0.870, 0.589 to 0.762, and 0.101 to 0.399, respectively. Tokenizer fertility follows the inverse pattern, averaging approximately 1.14 for English, 1.78 to 2.03 for Hausa, and 2.57 to 2.95 for Yoruba. Lower agreement thus aligns with higher subword fertility in this comparison. However, because these observations originate from only three languages, evaluating multiple models on the same language does not yield independent cross-lingual samples.

Sentiment remains predictable despite these lower agreement scores. Across all 24 combinations of model, language, and representation method, every separately selected probe achieves an interval lower bound above random chance (0.50). This pattern holds at the layer selected for direction reliability across all 23 eligible combinations. For instance, Gemma 4 E2B on Yoruba yields a final token agreement of 0.101 alongside a classification accuracy of 0.625; under mean pooling, agreement reaches 0.317 while accuracy rises to 0.692. A classifier can therefore reliably predict labels even when the underlying extracted direction varies across samples. Because accuracy and cosine similarity operate on distinct scales, where random chance is 0.50 and orthogonal directions yield zero, their absolute difference does not quantify predictability relative to reliability. Instead, the key takeaway is that prediction remains viable at layers with weak direction agreement.

Representation extraction methods further influence consistency. Final token agreement exceeds mean pooling agreement for every English and Hausa model under the primary topic division, with paired difference intervals excluding zero across all four English models and two Hausa models. On the other hand, Yoruba shows mixed directional trends with all three available intervals overlapping zero. The fourth Yoruba comparison is unavailable because no AfroLlama layer satisfies the 0.15 length overlap threshold under mean pooling. These findings demonstrate that mean pooling does not consistently outperform final token extraction for Yoruba, nor do they define a specific fertility threshold where pooling becomes advantageous.

\mainpage
\section{Sensitivity to selection and estimation choices}
\label{sec:robust}
\begin{table}[H]
\centering\small
\caption{Layer availability across twenty divisions of the topics at threshold 0.15. The language columns count divisions with an eligible layer. The final column counts English $>$ Hausa $>$ Yoruba among divisions with all three results. Unavailable comparisons are excluded from this count.}
\label{tab:ordering}
\begin{tabular}{llrrrr}
\toprule
Model & Pool & English & Hausa & Yoruba & Ordered / complete \\
\midrule
E2B & last & 20/20 & 20/20 & 18/20 & 18/18 \\
E2B & mean & 20/20 & 19/20 & 18/20 & 15/17 \\
E4B & last & 20/20 & 20/20 & 19/20 & 19/19 \\
E4B & mean & 20/20 & 19/20 & 19/20 & 16/18 \\
12B & last & 20/20 & 20/20 & 20/20 & 20/20 \\
12B & mean & 20/20 & 19/20 & 19/20 & 16/18 \\
AfroLlama & last & 20/20 & 20/20 & 20/20 & 20/20 \\
AfroLlama & mean & 20/20 & 18/20 & 14/20 & 9/12 \\
\bottomrule
\end{tabular}

\end{table}

We repeat layer selection and evaluation across twenty specified topic divisions using the same dataset, allocating ten topics for selection and fifteen for evaluation in each split. Because these splits divide the same underlying observations, including the primary division, they evaluate sensitivity to topic selection instead of replication on independent data.

The relative language ordering (English > Hausa > Yoruba) using the final token position remains perfectly consistent across all 77 valid comparisons (Table~\ref{tab:ordering}), with three comparisons having no eligible Yoruba layer. Under mean pooling, this ordering holds in 56 of 65 complete comparisons, alongside fifteen unavailable cases. Language rank ordering is therefore noticeably more consistent when using final token representations.

While overall rank order remains steady, individual numerical values show substantial variation. For example, E2B Yoruba final token agreement ranges from $-$0.117 to 0.275 across eighteen eligible divisions. Similarly, for 12B English under mean pooling, the primary division agreement of 0.473 marks the minimum across all twenty splits, where the median reaches 0.717. For AfroLlama on Yoruba under mean pooling, fourteen divisions yield an eligible layer even though none qualifies in the primary division. These ranges illustrate observed empirical variability instead of statistical confidence bounds, which is why we present the primary division alongside this sensitivity analysis.

Applying the sentence length constraint directly influences which layer outputs can be reported. Without this check, AfroLlama on Yoruba under mean pooling selects layer 2, yielding a selection agreement of 0.943 alongside a length overlap of 0.997. Because no layer satisfies the 0.15 overlap limit in that division, raising the threshold to 0.20 allows a layer to qualify. Evaluating thresholds of 0.10, 0.15, 0.20, and 0.25 reveals that sixteen of the 24 combinations maintain the same eligible layer, whereas eight experience changes in layer choice or availability. This threshold directly shapes model selection and should be reported alongside the primary findings.

Finally, we compare direction vectors derived from mean difference representations against the weight vectors of logistic probes. Each probe is trained on one half of the evaluation topics, with its parameters transformed back into the original representation space for comparison. Both techniques evaluate the same layer across identical topic divisions. Mean difference vectors yield higher split-half agreement across all 23 eligible combinations, with paired difference intervals excluding zero in seventeen cases. While this finding holds for our specific probe parameters and layer selection protocol, it does not demonstrate that mean differences are universally superior across all extraction settings. Complete results appear in Appendix~\ref{app:matched}.

\mainpage
\section{Comparisons between text sources}
\label{sec:provenance}
We compare each translated arm with native writing at the layer selected from native data, using the same fifteen evaluation topics. Across fifteen eligible combinations for Hausa and Yoruba, median cosine similarity is 0.338 for human translation, 0.398 for machine translation and 0.756 for translation into English and back. Human translation has the lowest agreement within an arm in fourteen combinations. This pattern describes the present dataset and does not establish a ranking of translation quality.

Texts translated into English and back remain closest to native writing across all fifteen combinations. Because these pairs originate directly from native sentences, this similarity does not provide independent confirmation of the extracted direction. We also report an attenuation reference score $h_{AX}$, representing expected agreement under classical measurement error assumptions:
\begin{equation}
 h_{AX}=\sqrt{g(r_A)g(r_X)},\qquad g(r)=\frac{2r}{1+r}.
\label{eq:ceiling}
\end{equation}
The function $g$ applies the Spearman--Brown correction for sample halves \citep{brown1910,spearman1910}, while $h$ uses the standard attenuation relation \citep{spearman1904}. These underlying assumptions do not necessarily hold for this representations, and we leave the reference undefined when reliability is negative. All fifteen comparisons with back-translated texts exceed this reference score, though shared topic content across text sources may introduce dependent errors in human and machine translations as well. Appendix~\ref{app:cross} reports complete cosine similarities, attenuation reference values, and confidence intervals. This reference score is neither a strict upper bound nor a test of statistical equivalence.

\section{Discussion}
Split-half agreement exposes sample sensitivity even when a probe predicts sentiment accurately. Successful probe performance does not demonstrate that the direction used to represent sentiment is reproducible. Distinct topic samples can contain different predictive features, enabling accurate classification even when estimated direction vectors vary. Similarly, the lower agreement observed among probe coefficient vectors suggests that successful classification can occur across multiple decision boundaries. These mechanisms represent potential explanatory factors that our experimental design does not evaluate directly.

High direction agreement alone remains insufficient for representation analysis. As demonstrated by the sentence length checks, a consistently estimated vector can reflect surface properties of the text instead of sentiment. Researchers evaluating representation geometry should assess classification performance alongside direction agreement, inspect identifiable influences such as sentence length, and select layers using topic sets distinct from evaluation topics. This approach adds a reliability check to existing frameworks for probe interpretation \citep{hewitt2019control,belinkov2022probing}.
Several scope constraints frame these empirical findings. This study evaluates a single sentiment contrast type across three languages. Three native authors per language completed distinct topic blocks, meaning that topic partitioning does not separate topic choice from writer-specific effects. Furthermore, evaluation relies on a dataset containing fifteen topics per split, yielding approximate confidence intervals. Subword fertility also covaries with pretraining exposure, language family, and specific structural features; as a result, neither tokenizer comparisons nor model comparisons isolate an explicit causal mechanism. 

\section{Conclusion}
Sentiment remains predictable across these datasets even when extracted directions show weak agreement across samples. Split-half agreement calculated from final token representations is consistently highest for English, followed by Hausa and Yoruba, across all valid comparisons in the tested topic divisions. Specific numerical values and layer availability exhibit greater variability, particularly when using mean pooling.
Measuring split-half agreement provides a direct assessment of direction reproducibility alongside classification performance. To ensure reproducibility in representation interpretability studies, agreement metrics should be reported with explicit details regarding topic partitioning, layer selection rules, and sentence length constraints. Determining whether tokenizer characteristics directly cause these cross-lingual variations will require controlled experimental designs.
\label{main-end}
\clearpage
\bibliographystyle{plainnat}
\bibliography{references}
\clearpage
\appendix
\raggedbottom
\section{Extraction and evaluation protocol}
\label{app:protocol}
We use Arm A for primary direction agreement and probe classification experiments, and Arms B, C, and D for comparisons across text sources. Text sources are aligned by topic; independently authored pairs do not share identical identifier strings. Three native authors per language authored distinct topic blocks. Because source metadata contains incomplete writer tracking tags, we could not evaluate generalization across separate authors. We used \texttt{nllb-200-3.3B} \citep{costajussa2022nllb} to generate Arms C and D.

Each normalized sequence was passed directly to the model tokenizer without chat formatting or prompt wrappers. We retained special tokens and applied right-side padding. Final-token extraction uses the final index where the attention mask equals one; mean pooling averages all sequence positions where the attention mask equals one, including special tokens.

Layer index zero corresponds to the embedding layer output. Models E2B and E4B are treated as distinct architectures instead of nested or identically trained variants. Exact model checkpoints, package dependencies, and random seed specifications appear in the reproducibility code repository.

The maximum sequence length reaches 68 tokens for Gemma architectures and 79 tokens for AfroLlama; no text required truncation. Subword fertility represents the aggregate token count excluding special tokens divided by total whitespace word count across both sides of all native pairs, instead of an average of individual sentence ratios.

We set $0.15$ as the primary length overlap threshold, using $0.10$, $0.20$, and $0.25$ for sensitivity checks. Analysis code is available at \url{https://github.com/mohdasaid/provenance-repe}. The repository contains evaluation scripts and plotting code.

\section{Uncertainty and numerical checks}\label{app:uncertainty}For a statistic $T$ computed from $m=15$ evaluation topics, let $T_{(-j)}$ denote its value after excluding all four pairs associated with topic $j$. Each deletion removes the topic from the existing divisions without drawing new divisions. The selected layer remains fixed throughout this procedure. Standard errors are estimated using the jackknife framework described by \citet{shalizi2023jackknife} to measure variation in the statistic across topic samples. Interval bounds are constructed using a Student $t$ multiplier:
\begin{equation}\widehat{\mathrm{SE}}(T)=\left[\frac{m-1}{m}\sum_{j=1}^{m}\bigl(T_{(-j)}-\overline{T}{(-\cdot)}\bigr)^2\right]^{1/2},
\qquad T\pm t{m-1,0.975}\widehat{\mathrm{SE}}(T).\end{equation}
Here, $\overline{T}_{(-\cdot)}$ is the mean over the fifteen topic deletions, and $t_{m-1,0.975}$ denotes the 97.5th percentile of the Student $t$ distribution with fourteen degrees of freedom. Cosine intervals are clipped to $[-1,1]$, paired reliability differences to $[-2,2]$, and intervals for the attenuation reference to $[0,1]$. Negative reliability values are not clipped to zero prior to applying the attenuation transformation. If an evaluation step yields an undefined output following a topic deletion, its corresponding interval is left reported as unavailable. We construct the interval by removing each topic in turn and recalculating the difference between the attenuation reference and observed cosine similarity.

These intervals approximate measurement uncertainty in extraction statistics for the observed sample sizes under the specified protocol. They remain strictly conditional on the selected layer, selection sample, and topic partitioning schedule. Consequently, they do not estimate uncertainty originating from new authors or repeated layer selection, nor do they demonstrate the validity of the underlying attenuation model.

Probe classification accuracy evaluates mean correctness across the two messages in each pair and subsequently across evaluation pairs. Interval estimation uses a non-parametric bootstrap resampling fifteen topics with replacement across 1,000 iterations, carrying all paired scores for a sampled topic together. The resulting interval spans the 2.5th to 97.5th percentiles of the bootstrapped accuracy distribution. Classifier parameters and layer selection remain fixed. In datasets yielding perfect classification, every resample maintains identical performance, producing an interval with identical lower and upper bounds. Such an interval does not prove perfect general accuracy.

For the matched estimator comparison, both extraction procedures evaluate the first twenty topic divisions. For each topic deletion, both probe halves undergo complete retraining, with standardization parameters recomputed strictly on the remaining training sequences. The confidence interval is calculated from the paired difference between mean-difference agreement and probe-coefficient agreement under identical topic deletions. These intervals are approximate and unadjusted; their empirical coverage has not been independently validated for probe weight vectors.

Interval behavior for direction agreement and cosine similarity was verified through synthetic simulations across seven scenario configurations using 16, 64, and 1,536 dimensions, encompassing weak and strong signals, unequal coordinate variances, heavy-tailed Student-$t_5$ noise, and correlated noise conditions. Each scenario evaluated 300 synthetic datasets alongside 10,000 reference iterations. Across checks of reliability, cosine similarity, and pooling method differences, empirical coverage ranged from 93.3\% to 99.7\%. This finite simulation result provides empirical context instead of a guaranteed coverage bound for the primary text collection, probe weight comparisons, or attenuation reference scores.

Numerical calculations were validated independently using a parallel codebase operating on the saved activation tensors. The accompanying reproducibility repository details these numerical checks and their operational bounds.

\section{Native results and probe evaluation}
\label{app:native}
Tables~\ref{tab:lastfull} and \ref{tab:meanfull} report reliability and decoding at the same layer. All intervals in this appendix are approximate conditional 95\% intervals. A dash means the relevant layer or transformed quantity is unavailable, not that it equals zero.

\begin{table}[H]
\centering\small
\caption{Native final token results at the layer selected for reliability.}
\label{tab:lastfull}
\begin{tabular}{llrll}
\toprule
Model & Language & Layer & $r$ [95\% interval] & Accuracy [95\% interval] \\
\midrule
E2B & English & 25 & 0.852 [0.792, 0.911] & 0.983 [0.958, 1.000] \\
E2B & Hausa & 16 & 0.703 [0.606, 0.800] & 0.767 [0.683, 0.842] \\
E2B & Yoruba & 16 & 0.101 [-0.100, 0.302] & 0.625 [0.542, 0.717] \\
E4B & English & 25 & 0.870 [0.827, 0.914] & 1.000 [1.000, 1.000] \\
E4B & Hausa & 25 & 0.762 [0.676, 0.848] & 0.808 [0.742, 0.875] \\
E4B & Yoruba & 26 & 0.399 [0.207, 0.591] & 0.683 [0.608, 0.758] \\
12B & English & 34 & 0.804 [0.742, 0.867] & 0.992 [0.975, 1.000] \\
12B & Hausa & 29 & 0.598 [0.462, 0.734] & 0.808 [0.758, 0.858] \\
12B & Yoruba & 29 & 0.185 [0.008, 0.361] & 0.625 [0.567, 0.700] \\
AfroLlama & English & 19 & 0.737 [0.692, 0.781] & 0.967 [0.933, 0.992] \\
AfroLlama & Hausa & 16 & 0.589 [0.551, 0.627] & 0.867 [0.817, 0.909] \\
AfroLlama & Yoruba & 27 & 0.329 [0.159, 0.499] & 0.742 [0.667, 0.825] \\
\bottomrule
\end{tabular}

\end{table}
\begin{table}[H]
\centering\small
\caption{Native mean pooling results at the layer selected for reliability. AfroLlama Yoruba has no eligible layer at threshold 0.15.}
\label{tab:meanfull}
\begin{tabular}{llrll}
\toprule
Model & Language & Layer & $r$ [95\% interval] & Accuracy [95\% interval] \\
\midrule
E2B & English & 12 & 0.723 [0.660, 0.787] & 0.942 [0.900, 0.975] \\
E2B & Hausa & 9 & 0.424 [0.268, 0.580] & 0.733 [0.650, 0.817] \\
E2B & Yoruba & 22 & 0.317 [0.045, 0.588] & 0.692 [0.617, 0.767] \\
E4B & English & 19 & 0.789 [0.723, 0.855] & 0.975 [0.950, 1.000] \\
E4B & Hausa & 18 & 0.609 [0.489, 0.728] & 0.867 [0.817, 0.917] \\
E4B & Yoruba & 34 & 0.271 [-0.174, 0.716] & 0.642 [0.583, 0.708] \\
12B & English & 3 & 0.473 [0.180, 0.767] & 0.750 [0.633, 0.850] \\
12B & Hausa & 16 & 0.531 [0.343, 0.719] & 0.733 [0.658, 0.825] \\
12B & Yoruba & 32 & 0.347 [0.075, 0.619] & 0.667 [0.600, 0.733] \\
AfroLlama & English & 32 & 0.463 [0.361, 0.565] & 0.917 [0.875, 0.950] \\
AfroLlama & Hausa & 0 & 0.430 [0.261, 0.600] & 0.700 [0.617, 0.783] \\
AfroLlama & Yoruba & -- & -- & -- \\
\bottomrule
\end{tabular}

\end{table}

\begin{table}[H]
\centering\small
\caption{Accuracy after a separate search for a probe layer within selection data. These are distinct from the results at the same layer above. Fertility is computed on native text.}
\label{tab:probefull}
\begin{tabular}{llrll}
\toprule
Model & Language & Fertility & Final token accuracy & Mean-pool accuracy \\
\midrule
E2B & English & 1.137 & 0.942 [0.900, 0.975] & 0.925 [0.883, 0.967] \\
E2B & Hausa & 1.783 & 0.775 [0.692, 0.858] & 0.783 [0.708, 0.867] \\
E2B & Yoruba & 2.572 & 0.592 [0.533, 0.642] & 0.642 [0.592, 0.692] \\
E4B & English & 1.137 & 0.975 [0.950, 1.000] & 0.958 [0.917, 0.992] \\
E4B & Hausa & 1.783 & 0.833 [0.775, 0.892] & 0.808 [0.733, 0.883] \\
E4B & Yoruba & 2.572 & 0.692 [0.608, 0.767] & 0.642 [0.567, 0.717] \\
12B & English & 1.137 & 0.992 [0.975, 1.000] & 0.967 [0.942, 0.992] \\
12B & Hausa & 1.783 & 0.808 [0.758, 0.858] & 0.667 [0.575, 0.758] \\
12B & Yoruba & 2.572 & 0.625 [0.567, 0.700] & 0.633 [0.558, 0.700] \\
AfroLlama & English & 1.139 & 0.967 [0.942, 0.992] & 0.942 [0.908, 0.975] \\
AfroLlama & Hausa & 2.034 & 0.900 [0.858, 0.942] & 0.867 [0.800, 0.925] \\
AfroLlama & Yoruba & 2.948 & 0.725 [0.658, 0.792] & 0.833 [0.783, 0.892] \\
\bottomrule
\end{tabular}

\end{table}
There are 120 evaluation messages per probe. Since each pair contains one positive and one negative label, chance accuracy is 0.5 and ordinary accuracy equals balanced accuracy. These intervals use the fixed classifiers and topic bootstrap described in Appendix~\ref{app:uncertainty}.

\section{Readout and threshold sensitivity}
\label{app:thresholds}
\begin{table}[H]
\centering\small
\caption{Paired difference between final token and mean pooling reliability in the primary allocation. The two readouts may select different layers. Intervals delete the same topic in both conditions.}
\begin{tabular}{lll}
\toprule
Model & Language & $r_{\mathrm{last}}-r_{\mathrm{mean}}$ [95\% interval] \\
\midrule
E2B & English & 0.129 [0.089, 0.168] \\
E2B & Hausa & 0.279 [0.092, 0.466] \\
E2B & Yoruba & -0.216 [-0.483, 0.052] \\
E4B & English & 0.081 [0.051, 0.111] \\
E4B & Hausa & 0.154 [0.030, 0.278] \\
E4B & Yoruba & 0.128 [-0.378, 0.634] \\
12B & English & 0.331 [0.057, 0.605] \\
12B & Hausa & 0.067 [-0.195, 0.329] \\
12B & Yoruba & -0.162 [-0.420, 0.096] \\
AfroLlama & English & 0.274 [0.200, 0.347] \\
AfroLlama & Hausa & 0.159 [-0.025, 0.342] \\
AfroLlama & Yoruba & -- \\
\bottomrule
\end{tabular}

\end{table}
\begin{table}[H]
\centering\small
\caption{Selected layer and reliability on evaluation data in parentheses at each threshold for length overlap. A dash indicates no eligible layer. Sixteen configurations retain the same eligible layer across all four thresholds.}
\begin{tabular}{lllllll}
\toprule
Model & Lang. & Pool & 0.10 & 0.15 & 0.20 & 0.25 \\
\midrule
E2B & eng & last & 25 (0.852) & 25 (0.852) & 25 (0.852) & 25 (0.852) \\
E2B & eng & mean & 12 (0.723) & 12 (0.723) & 12 (0.723) & 12 (0.723) \\
E2B & hau & last & 16 (0.703) & 16 (0.703) & 16 (0.703) & 16 (0.703) \\
E2B & hau & mean & 9 (0.424) & 9 (0.424) & 9 (0.424) & 9 (0.424) \\
E2B & yor & last & 16 (0.101) & 16 (0.101) & 16 (0.101) & 16 (0.101) \\
E2B & yor & mean & 23 (0.330) & 22 (0.317) & 22 (0.317) & 22 (0.317) \\
E4B & eng & last & 25 (0.870) & 25 (0.870) & 25 (0.870) & 25 (0.870) \\
E4B & eng & mean & 19 (0.789) & 19 (0.789) & 19 (0.789) & 19 (0.789) \\
E4B & hau & last & 25 (0.762) & 25 (0.762) & 25 (0.762) & 25 (0.762) \\
E4B & hau & mean & 18 (0.609) & 18 (0.609) & 18 (0.609) & 18 (0.609) \\
E4B & yor & last & 26 (0.399) & 26 (0.399) & 24 (0.370) & 24 (0.370) \\
E4B & yor & mean & 34 (0.271) & 34 (0.271) & 34 (0.271) & 34 (0.271) \\
12B & eng & last & 34 (0.804) & 34 (0.804) & 34 (0.804) & 34 (0.804) \\
12B & eng & mean & 3 (0.473) & 3 (0.473) & 3 (0.473) & 3 (0.473) \\
12B & hau & last & 29 (0.598) & 29 (0.598) & 29 (0.598) & 23 (0.499) \\
12B & hau & mean & 16 (0.531) & 16 (0.531) & 16 (0.531) & 45 (0.557) \\
12B & yor & last & 29 (0.185) & 29 (0.185) & 29 (0.185) & 29 (0.185) \\
12B & yor & mean & 37 (0.333) & 32 (0.347) & 32 (0.347) & 32 (0.347) \\
AfroLlama & eng & last & 19 (0.737) & 19 (0.737) & 19 (0.737) & 19 (0.737) \\
AfroLlama & eng & mean & 32 (0.463) & 32 (0.463) & 32 (0.463) & 32 (0.463) \\
AfroLlama & hau & last & 16 (0.589) & 16 (0.589) & 16 (0.589) & 16 (0.589) \\
AfroLlama & hau & mean & 0 (0.430) & 0 (0.430) & 31 (0.482) & 29 (0.486) \\
AfroLlama & yor & last & -- & 27 (0.329) & 14 (0.231) & 14 (0.231) \\
AfroLlama & yor & mean & -- & -- & 32 (0.385) & 32 (0.385) \\
\bottomrule
\end{tabular}

\end{table}

\section{Repeated topic allocations}
\label{app:allocations}
We summarise all twenty topic allocations below. Minima and maxima describe variation across eligible allocations; they are not confidence limits. The allocations reuse the same data.
\begin{table}[H]
\centering\small
\caption{Agreement across twenty allocations at threshold 0.15. Median and range use eligible allocations only. The final column counts distinct selected layers.}
\begin{tabular}{lllrrlr}
\toprule
Model & Lang. & Pool & Eligible & Median & Range & Layers \\
\midrule
E2B & eng & last & 20/20 & 0.864 & [0.826, 0.892] & 5 \\
E2B & eng & mean & 20/20 & 0.753 & [0.716, 0.798] & 4 \\
E2B & hau & last & 20/20 & 0.563 & [0.309, 0.703] & 6 \\
E2B & hau & mean & 19/20 & 0.432 & [0.255, 0.524] & 9 \\
E2B & yor & last & 18/20 & 0.099 & [-0.117, 0.275] & 12 \\
E2B & yor & mean & 18/20 & 0.208 & [-0.067, 0.458] & 10 \\
E4B & eng & last & 20/20 & 0.884 & [0.841, 0.905] & 4 \\
E4B & eng & mean & 20/20 & 0.814 & [0.783, 0.836] & 2 \\
E4B & hau & last & 20/20 & 0.659 & [0.496, 0.762] & 3 \\
E4B & hau & mean & 19/20 & 0.592 & [0.403, 0.658] & 9 \\
E4B & yor & last & 19/20 & 0.387 & [-0.021, 0.508] & 6 \\
E4B & yor & mean & 19/20 & 0.399 & [0.166, 0.575] & 13 \\
12B & eng & last & 20/20 & 0.830 & [0.792, 0.853] & 4 \\
12B & eng & mean & 20/20 & 0.717 & [0.473, 0.771] & 9 \\
12B & hau & last & 20/20 & 0.601 & [0.427, 0.658] & 9 \\
12B & hau & mean & 19/20 & 0.494 & [0.432, 0.638] & 8 \\
12B & yor & last & 20/20 & 0.157 & [-0.087, 0.292] & 13 \\
12B & yor & mean & 19/20 & 0.304 & [-0.071, 0.644] & 12 \\
AfroLlama & eng & last & 20/20 & 0.747 & [0.720, 0.779] & 4 \\
AfroLlama & eng & mean & 20/20 & 0.478 & [0.184, 0.505] & 3 \\
AfroLlama & hau & last & 20/20 & 0.484 & [0.376, 0.589] & 5 \\
AfroLlama & hau & mean & 18/20 & 0.408 & [0.304, 0.467] & 8 \\
AfroLlama & yor & last & 20/20 & 0.284 & [0.066, 0.358] & 10 \\
AfroLlama & yor & mean & 14/20 & 0.322 & [0.137, 0.480] & 3 \\
\bottomrule
\end{tabular}

\end{table}
\ifextended\else
\begin{figure}[H]
\centering
\includegraphics[width=\linewidth]{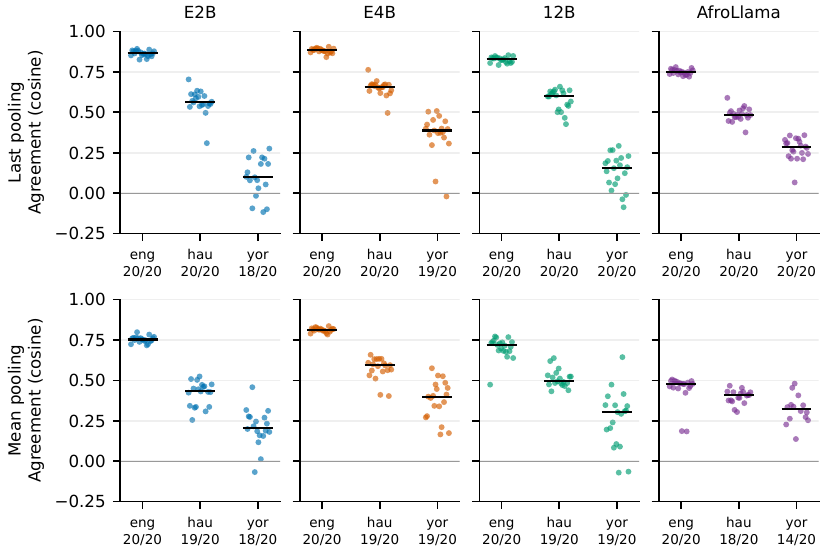}
\caption{Agreement across topic allocations. Each dot is an eligible allocation and the horizontal mark is its configuration's median. Counts show eligibility out of twenty.}
\end{figure}
\fi

\section{Matched direction estimators}
\label{app:matched}
Both estimators use the same twenty topic divisions into halves and the same layer selected from native data. Logistic coefficients are divided by the training scaler's feature standard deviations before the cosine is computed. The intercept is omitted. Directions from mean differences are not standardised. Both fitting procedures are therefore evaluated in the original activation coordinates, although the probe's regularisation still depends on its training standardisation.
\begin{table}[H]
\centering\small
\caption{Agreement of mean difference and probe coefficient vectors. The difference is paired by topic deletion. Seventeen unadjusted approximate intervals exclude zero. These comparisons use twenty partitions, so their values for mean differences need not equal the main estimates based on 100 divisions.}
\begin{tabular}{lllrrl}
\toprule
Model & Lang. & Pool & $r_{\mathrm{mean\ difference}}$ & $r_{\mathrm{probe\ weights}}$ & Difference [95\% interval] \\
\midrule
E2B & eng & last & 0.858 & 0.516 & 0.342 [0.243, 0.442] \\
E2B & eng & mean & 0.724 & 0.182 & 0.542 [0.152, 0.932] \\
E2B & hau & last & 0.708 & 0.298 & 0.410 [0.071, 0.750] \\
E2B & hau & mean & 0.399 & 0.254 & 0.145 [-0.380, 0.671] \\
E2B & yor & last & 0.105 & -0.016 & 0.121 [-0.272, 0.513] \\
E2B & yor & mean & 0.315 & -0.039 & 0.354 [-0.088, 0.796] \\
E4B & eng & last & 0.877 & 0.676 & 0.201 [0.153, 0.248] \\
E4B & eng & mean & 0.790 & 0.327 & 0.463 [0.307, 0.620] \\
E4B & hau & last & 0.764 & 0.292 & 0.472 [0.236, 0.708] \\
E4B & hau & mean & 0.612 & 0.112 & 0.500 [0.193, 0.807] \\
E4B & yor & last & 0.432 & 0.180 & 0.253 [0.007, 0.498] \\
E4B & yor & mean & 0.211 & 0.116 & 0.096 [-0.403, 0.594] \\
12B & eng & last & 0.812 & 0.397 & 0.415 [0.183, 0.647] \\
12B & eng & mean & 0.477 & 0.136 & 0.341 [0.059, 0.623] \\
12B & hau & last & 0.609 & 0.148 & 0.461 [0.167, 0.756] \\
12B & hau & mean & 0.517 & 0.141 & 0.376 [0.230, 0.522] \\
12B & yor & last & 0.216 & 0.116 & 0.100 [-0.185, 0.385] \\
12B & yor & mean & 0.318 & 0.155 & 0.162 [-0.126, 0.451] \\
AfroLlama & eng & last & 0.741 & 0.481 & 0.260 [0.182, 0.338] \\
AfroLlama & eng & mean & 0.460 & 0.371 & 0.089 [0.021, 0.157] \\
AfroLlama & hau & last & 0.590 & 0.332 & 0.258 [0.151, 0.365] \\
AfroLlama & hau & mean & 0.425 & 0.218 & 0.206 [0.043, 0.370] \\
AfroLlama & yor & last & 0.365 & 0.133 & 0.232 [0.045, 0.420] \\
\bottomrule
\end{tabular}

\end{table}
We selected layers to maximise agreement of mean differences on separate topics. We did not repeat the comparison with layers selected for probe stability, so the estimator ranking remains conditional on this choice and the fixed regularisation. Appendix~\ref{app:uncertainty} describes the interval limitations.

\section{Provenance comparisons}
\label{app:cross}
All comparisons use the layer selected from native data and the same evaluation topic labels. Agreement within each arm is shown first. The following pages report observed cosine between native writing and another arm $c$, attenuation reference $h$ from Equation~\ref{eq:ceiling}, and their paired difference. The reference assumes a relationship between split half agreement and reliability for the full sample that is not established for these representations. Shared topics and derived text can also induce dependent errors.
\begin{table}[H]
\centering\small
\caption{Agreement within each arm for native writing (A), human translation (B), machine translation (C) and translation into English and back (D). AfroLlama Yoruba mean pooling is unavailable. Full arm intervals are included in the accompanying CSV tables.}
\begin{tabular}{lllrrrr}
\toprule
Model & Lang. & Pool & A & B & C & D \\
\midrule
E2B & hau & last & 0.703 & 0.332 & 0.565 & 0.670 \\
E2B & hau & mean & 0.424 & 0.347 & 0.403 & 0.520 \\
E2B & yor & last & 0.101 & 0.019 & 0.222 & 0.242 \\
E2B & yor & mean & 0.317 & 0.009 & 0.325 & 0.269 \\
E4B & hau & last & 0.762 & 0.457 & 0.633 & 0.734 \\
E4B & hau & mean & 0.609 & 0.462 & 0.537 & 0.649 \\
E4B & yor & last & 0.399 & 0.153 & 0.569 & 0.287 \\
E4B & yor & mean & 0.271 & -0.053 & 0.438 & 0.335 \\
12B & hau & last & 0.598 & 0.362 & 0.495 & 0.588 \\
12B & hau & mean & 0.531 & 0.360 & 0.415 & 0.508 \\
12B & yor & last & 0.185 & 0.152 & 0.453 & 0.156 \\
12B & yor & mean & 0.347 & 0.236 & 0.307 & 0.217 \\
AfroLlama & hau & last & 0.589 & 0.399 & 0.530 & 0.569 \\
AfroLlama & hau & mean & 0.430 & 0.147 & 0.217 & 0.518 \\
AfroLlama & yor & last & 0.329 & 0.168 & 0.487 & 0.292 \\
\bottomrule
\end{tabular}

\end{table}
\begin{table}[H]
\centering\footnotesize
\setlength{\tabcolsep}{3pt}
\caption{Native versus human translation (A--B). Values are followed by full conditional 95\% intervals. An interval marked [--] is undefined because a required reliability after a topic is removed is negative. The E4B Yoruba mean pooling reference is itself undefined because the point reliability for B is negative.}
\begin{tabular}{llllll}
\toprule
Model & Lang. & Pool & $c$ [95\% interval] & $h$ [95\% interval] & $h-c$ [95\% interval] \\
\midrule
E2B & hau & last & 0.464 [-0.042, 0.970] & 0.641 [0.527, 0.756] & 0.178 [-0.348, 0.703] \\
E2B & hau & mean & 0.476 [0.319, 0.634] & 0.554 [0.387, 0.721] & 0.078 [-0.055, 0.210] \\
E2B & yor & last & 0.058 [-0.160, 0.276] & 0.082 [--] & 0.024 [--] \\
E2B & yor & mean & 0.143 [-0.370, 0.657] & 0.091 [--] & -0.052 [--] \\
E4B & hau & last & 0.429 [-0.018, 0.877] & 0.737 [0.645, 0.829] & 0.307 [-0.175, 0.789] \\
E4B & hau & mean & 0.616 [0.490, 0.741] & 0.692 [0.593, 0.790] & 0.076 [-0.058, 0.210] \\
E4B & yor & last & 0.327 [0.100, 0.553] & 0.389 [0.104, 0.674] & 0.062 [-0.289, 0.414] \\
E4B & yor & mean & 0.080 [-0.831, 0.990] & -- & -- \\
12B & hau & last & 0.362 [0.034, 0.690] & 0.631 [0.541, 0.720] & 0.269 [-0.119, 0.656] \\
12B & hau & mean & 0.456 [0.220, 0.693] & 0.606 [0.468, 0.744] & 0.150 [-0.002, 0.301] \\
12B & yor & last & 0.240 [0.001, 0.479] & 0.287 [0.120, 0.454] & 0.047 [-0.147, 0.241] \\
12B & yor & mean & 0.211 [-0.154, 0.576] & 0.443 [0.293, 0.594] & 0.232 [-0.151, 0.615] \\
AfroLlama & hau & last & 0.465 [0.249, 0.682] & 0.650 [0.616, 0.685] & 0.185 [-0.034, 0.404] \\
AfroLlama & hau & mean & 0.338 [0.093, 0.583] & 0.393 [0.190, 0.596] & 0.055 [-0.223, 0.333] \\
AfroLlama & yor & last & 0.285 [0.148, 0.422] & 0.377 [0.220, 0.535] & 0.092 [-0.084, 0.269] \\
\bottomrule
\end{tabular}

\end{table}

\begin{table}[H]
\centering\footnotesize
\setlength{\tabcolsep}{3pt}
\caption{Native versus machine translation (A--C). The text is translated from English seeds. Independent errors relative to native writing are not established by this construction because the conditions share topics and evaluation choices.}
\begin{tabular}{llllll}
\toprule
Model & Lang. & Pool & $c$ [95\% interval] & $h$ [95\% interval] & $h-c$ [95\% interval] \\
\midrule
E2B & hau & last & 0.524 [0.335, 0.713] & 0.772 [0.693, 0.851] & 0.248 [0.042, 0.454] \\
E2B & hau & mean & 0.481 [0.211, 0.751] & 0.585 [0.456, 0.714] & 0.104 [-0.121, 0.328] \\
E2B & yor & last & 0.296 [0.100, 0.492] & 0.258 [0.000, 0.534] & -0.038 [-0.352, 0.276] \\
E2B & yor & mean & 0.473 [0.264, 0.681] & 0.486 [0.266, 0.705] & 0.013 [-0.260, 0.286] \\
E4B & hau & last & 0.576 [0.389, 0.763] & 0.819 [0.755, 0.883] & 0.243 [0.039, 0.447] \\
E4B & hau & mean & 0.660 [0.546, 0.774] & 0.727 [0.641, 0.813] & 0.067 [-0.038, 0.173] \\
E4B & yor & last & 0.364 [0.201, 0.528] & 0.643 [0.510, 0.776] & 0.279 [0.086, 0.471] \\
E4B & yor & mean & 0.398 [-0.006, 0.803] & 0.510 [0.132, 0.887] & 0.111 [-0.105, 0.327] \\
12B & hau & last & 0.375 [0.179, 0.570] & 0.704 [0.616, 0.792] & 0.329 [0.063, 0.596] \\
12B & hau & mean & 0.475 [0.230, 0.720] & 0.638 [0.497, 0.779] & 0.162 [-0.036, 0.361] \\
12B & yor & last & 0.291 [0.214, 0.368] & 0.441 [0.277, 0.606] & 0.150 [-0.008, 0.309] \\
12B & yor & mean & 0.221 [-0.024, 0.466] & 0.492 [0.301, 0.683] & 0.271 [0.014, 0.528] \\
AfroLlama & hau & last & 0.550 [0.426, 0.674] & 0.717 [0.680, 0.754] & 0.167 [0.024, 0.310] \\
AfroLlama & hau & mean & 0.382 [0.119, 0.645] & 0.463 [0.264, 0.662] & 0.080 [-0.226, 0.387] \\
AfroLlama & yor & last & 0.158 [0.055, 0.260] & 0.570 [0.438, 0.702] & 0.412 [0.217, 0.606] \\
\bottomrule
\end{tabular}

\end{table}
\begin{table}[H]
\centering\footnotesize
\setlength{\tabcolsep}{3pt}
\caption{Native versus translation into English and back (A--D). D is derived from A. All fifteen observed cosines exceed the attenuation reference, which must not be interpreted as a hard bound or as evidence of independent agreement.}
\begin{tabular}{llllll}
\toprule
Model & Lang. & Pool & $c$ [95\% interval] & $h$ [95\% interval] & $h-c$ [95\% interval] \\
\midrule
E2B & hau & last & 0.889 [0.826, 0.951] & 0.814 [0.750, 0.878] & -0.075 [-0.148, -0.001] \\
E2B & hau & mean & 0.807 [0.688, 0.925] & 0.638 [0.511, 0.766] & -0.168 [-0.254, -0.083] \\
E2B & yor & last & 0.437 [-0.001, 0.875] & 0.267 [0.000, 0.556] & -0.170 [-0.633, 0.293] \\
E2B & yor & mean & 0.696 [0.434, 0.958] & 0.451 [0.108, 0.794] & -0.244 [-0.458, -0.031] \\
E4B & hau & last & 0.926 [0.872, 0.979] & 0.856 [0.792, 0.920] & -0.070 [-0.132, -0.008] \\
E4B & hau & mean & 0.866 [0.777, 0.955] & 0.772 [0.699, 0.845] & -0.094 [-0.206, 0.018] \\
E4B & yor & last & 0.675 [0.457, 0.894] & 0.504 [0.272, 0.737] & -0.171 [-0.408, 0.067] \\
E4B & yor & mean & 0.756 [0.377, 1.000] & 0.463 [0.000, 1.000] & -0.293 [-0.657, 0.070] \\
12B & hau & last & 0.785 [0.682, 0.889] & 0.744 [0.664, 0.825] & -0.041 [-0.134, 0.052] \\
12B & hau & mean & 0.820 [0.702, 0.939] & 0.684 [0.539, 0.828] & -0.137 [-0.271, -0.003] \\
12B & yor & last & 0.517 [0.343, 0.691] & 0.290 [0.122, 0.458] & -0.227 [-0.410, -0.044] \\
12B & yor & mean & 0.729 [0.553, 0.905] & 0.429 [0.136, 0.722] & -0.300 [-0.488, -0.113] \\
AfroLlama & hau & last & 0.852 [0.809, 0.895] & 0.734 [0.694, 0.773] & -0.118 [-0.153, -0.084] \\
AfroLlama & hau & mean & 0.725 [0.624, 0.826] & 0.641 [0.495, 0.787] & -0.084 [-0.164, -0.004] \\
AfroLlama & yor & last & 0.625 [0.435, 0.814] & 0.473 [0.337, 0.610] & -0.151 [-0.390, 0.087] \\
\bottomrule
\end{tabular}

\end{table}
The medians reported in Section~\ref{sec:provenance} summarise different models and readouts. They are not independent estimates of a population translation effect.

\end{document}